\documentclass[a4paper,10pt, twocolumn]{article}
\usepackage{times}  
\usepackage{helvet}  
\usepackage{courier}  
\usepackage{url}  
\usepackage{graphicx}  
\usepackage{booktabs}
\usepackage{multirow}
\usepackage{amssymb}
\usepackage{subfig}
\usepackage{array}
\usepackage{fixltx2e}
\usepackage[version-1-compatibility]{siunitx}

\date{}

\newcommand{\figH}{2.2in}
\newcommand{\figHsmall}{2in}
\newcommand{\davide}{D.A.V.I.D.E.}

\begin{document}


\title{Anomaly Detection using Autoencoders in High Performance Computing Systems}
\author{Andrea Borghesi \\DEI, University of Bologna \and Andrea Bartolini \\DEI, University of Bologna \and Michele Lombardi \\DISI, University of Bologna \and Michela Milano \\DISI, University of Bologna \and Luca Benini \\Integrated Systems Laboratory, ETHZ}

\maketitle

\begin{abstract}
Anomaly detection in supercomputers is a very difficult problem due to the big scale of the systems and the high number of components. The current state of the art for automated anomaly detection employs Machine Learning methods or statistical regression models in a supervised fashion, meaning that the detection tool is trained to distinguish among a fixed set of behaviour classes (healthy and unhealthy states).

We propose a novel approach for anomaly detection in High Performance Computing systems based on a Machine (Deep) Learning technique, namely a type of neural network called \emph{autoencoder}. The key idea is to train a set of autoencoders to learn the normal (healthy) behaviour of the supercomputer nodes and, after training, use them to identify abnormal conditions. This is different from previous approaches which where based on learning the abnormal condition, for which there are much smaller datasets (since it is very hard to identify them to begin with). 

We test our approach on a real supercomputer equipped with a fine-grained, scalable monitoring infrastructure that can provide large amount of data to characterize the system behaviour. The results are extremely promising: after the training phase to learn the normal system behaviour, our method is capable of detecting anomalies that have never been seen before with a very good accuracy (values ranging between 88\% and 96\%). 
\end{abstract}

\section{Introduction}
\label{sec:intro}
High Performance Computing (HPC) systems are complex machines with many components that must operate concurrently at the best of their theoretical performance. In reality, many factors can degrade the performance of a HPC system: hardware can break, the applications may enter undesired and unexpected states, components can be wrongly configured. A critical aspect of modern and future supercomputers is the capability of detecting faulty conditions stemming from the improper behaviour of one or multiple parts. This issue is relevant not only for scientific computing systems but also in data centers and clouds providers, whose business strongly relies on the availability of their web services. For instance, Amazon in 2016 would have lost 15M\$ for just an hour of out of service \cite{hennessy2011computer}. An automated process for anomaly detection would be a great improvement for current HPC systems, and it will probably be a necessity for future Exascale supercomputers.

Nowadays, monitoring infrastructures are available in many HPC systems and data centers, used to gather data about the state of the systems and their components thanks to a large variety of measurement sensors. Given the deluge of data originating from a monitoring framework, real-time identification of problems and undesired situations is a daunting task for system administrators. The growing scale of HPC systems will only make this task even more difficult. In this paper we present a novel approach to deal with this issue, relying on a fine-grain monitoring framework and on an autonomous anomaly detection method that uses Machine Learning (ML) techniques. 

Automated anomaly detection is still a relatively unexplored area in the HPC field. The current state-of-the-art relies on \emph{supervised} \cite{mitchell1999machine} ML methods that learn to distinguish between healthy and faulty states after a training phase during which the supercomputer must be subjected to different conditions, namely the behaviour classes to be identified, for example normal behaviour and a set of anomalies. With this scheme, the anomaly detector learns to classify different classes using \emph{labeled} training data. 

This requirement complicates the training process: a supervised algorithm needs to be shown data containing examples of both healthy and unhealthy status (normal behaviour and anomalies). Moreover, the data set should ideally be unbiased and balanced, that is there should be roughly the same number of examples (data points in the set) for each class. In HPC systems, data is very abundant but labels are scarce. 

However, in supercomputers the normal behaviour is predominant -- and can be deterministically restored by system administrators. The same cannot be said for faulty behaviour, which is undesired, sporadic and uncontrolled. 
Furthermore, even when abnormal conditions are observed and dealt with, they are not necessarily stored in logging systems (unless the logging systems are explicitly designed for this), but rather the burden of assigning labels to a data set -- identify the corresponding class, healthy/unhealthy -- falls onto the system administrators, a less than ideal situation. Thus it is not easy and often neither possible to obtain the correct labeled data sets required by typical supervised approaches.

Conversely, there is another type of ML that does not require any label and it is referred to as \emph{unsupervised} \cite{mitchell1999machine} learning. In this case the data set contains only the features describing the system state and no labels; the learning algorithm learns useful properties about the structure of the dataset. 
To address the issue, we propose an anomaly detection method less dependent on labeled data; to be precise, our approach belongs to the \emph{semi-supervised} branch of ML, which combines the two methodologies described before. Our idea is to use autoencoders \cite{goodfellow2016deep} to learn the normal behaviour of supercomputer nodes and then to use them to detect abnormal states. In our method we require labels during the pre-processing phase because we need to obtain a data set containing only normal conditions. After this ``normal'' data set has been obtained the training of the ML model proceeds in unsupervised fashion, without the need of labels. A critical advantage of our method is that it will be able to identify faulty conditions even though these have not been encountered earlier during the training phase.
With our method we do not need to inject anomalies during the training phase (possibly not feasible in a production system) and we do not require system logs or changes to the standard supercomputer users' work flow.

The main contributions of our approach are: 1) a very precise anomaly detection rate (up to 88\%-96\% accuracy); 2) identification of new types of anomalies unseen during the initial training phase (thanks to its semi-supervised nature); 3) no need for large amount of labeled data. To demonstrate the feasibility of our approach we consider a real supercomputer hosted by the Italian inter-universities consortium CINECA \cite{CINECA}. We use historical data collected with an integrated monitoring system to train our autoencoders and then we test them by injecting anomalies in a subset of the computing nodes; the experimental results show how this approach can distinguish between normal and anomalous states with a very high level of accuracy.

\section{Related Works}
\label{sec:related}

\subsection{Anomaly Detection in HPC Systems}
\label{sec:related_HPC}

Tuncer et al. \cite{tuncer2017diagnosing} deal with the problem of diagnosing performance variations in HPC systems. The approach is based on the collection of several measurements gathered by a monitoring infrastructure; from these measures, a set of statistical features describing the state of the supercomputer is extracted. The authors then train different ML algorithms to classify the behaviour of the supercomputer using the statistical features previously mentioned. Unfortunately the authors propose a supervised approach which is not perfectly suited for the HPC context (as discussed previously).

Baseman et al. \cite{baseman2016interpretable} propose a similar method for anomaly detection in HPC systems. They apply a general statistical technique called \emph{classifier-adjusted density estimation} (CADE) to the HPC context. CADE relies on the observation that combining a uniform density estimate and the probabilistic output of a classifier results in an accurate density estimator. First they extract temporal relational features and their gradients from the sensor data. Then they use both real and artificially generated data (thanks to density estimation) to train a supervised classifier, specifically a  Random Forest classifier, in order to rank each data point depending on its ``anomalousness''. Again, the main limit of this work is to be based on a supervised approach.

Dani et al. \cite{dani2017k} present an unsupervised approach for anomaly detection in HPC. Their work is remarkably different from our approach since they do not rely on a monitoring infrastructure but consider only the console logs generated by computing nodes. The goal of the method is to distinguish log messages regarding faulty situations from logs generated by nodes in normal condition; in order to do so clustering methods (k-means) are used. This work focuses on faults that can be recognized by a node itself and recorded in a log message, thus greatly limiting the class of detectable anomalies. Conversely, in our approach we infer anomalies using simply the data collected by a monitoring infrastructure, without requiring a mechanism to identify anomalies on the nodes.

\subsection{Anomaly Detection with Deep Learning Approaches}
\label{sec:related_DL}
Although not yet applied to the HPC field, Deep Learning based approaches for anomaly detection have been studied in other areas \cite{kwon2017survey,kiran2018overview}, especially in recent years. 

Lv et al. \cite{lv2016fault} propose a deep learning based algorithm for fault diagnosis in chemical production systems. The proposed method is capable of real time  detection and classification and, moreover, it can do the diagnosis online. Nevertheless, their approach is supervised and thus it definitely differs from ours. Lee et al.  \cite{lee2017convolutional} introduce a convolutional neural network (CNN) model for fault identification and classification in semiconductor manufacturing processes. This method makes it possible to locate the variable and time information that represents process faults.

Costa et al. \cite{costa2015fully} describe a fully unsupervised algorithm for real-time detection of faults in industrial plants. The algorithm relies on the identification of a set of features that are then used to learn the normal behaviour of the plant, expressing it as a probability density estimation. The online classifier then uses the distance from the normal distribution to classify new data samples. The model is unsupervised and it can handle unseen types of anomalies. However, the approach is specifically targeted at plants for industrial process control and thus not well suited for our HPC system case. Particularly, a relatively small set of features is considered w.r.t. to the hundreds (thousands) of metrics found in a supercomputer.

Ince et al. \cite{ince2016real} discuss a CNN-based method for electrical motor fault detection; their method can work directly on the raw measurement data, with no preprocessing. The neural network combines feature extraction and classification, but proceeds in a supervised manner.

\section{Data Collection}
\label{sec:data_collection}
A very important aspect for our anomaly detection approach is the availability of large quantity of data that monitors and thus describes the state of a supercomputer. To test our approach we take advantage of a supercomputer with an integrated monitoring infrastructure able to handle large amounts of data coming from several different sources. Our target system is \davide \cite{DAVIDE}, an energy efficient supercomputer developed by E4 Computer Engineering \cite{e4} and hosted by CINECA in Bologna, Italy. It is composed by 45 nodes with a total peak performance of \SI{990}{\tera Flops} and an estimated power consumption of less than \SI{2}{\kilo \watt} per node. Each node hosts two IBM POWER8 Processors with NVIDIA NVLink and four Tesla P100 GPUs.
The system was ranked \#440 in TOP500 \cite{Top500} and \#18 in GREEN500 \cite{Green500} in November 2017 list.

The data collection infrastructure deployed in \davide~is called \emph{Examon} and has been presented in previous works \cite{beneventi2017continuous,DBLP:conf/cf/BartoliniBLBGTG18}. 
Examon is a fine-grained, lightweight and scalable monitoring infrastructure for Exascale supercomputers. The data coming from heterogeneous data sources is gathered in an integrated and uniform repository, making it very easy to create data sets providing a holistic view of the supercomputer and thus describing the system state.
The main components of the Examon framework are a set of agents running outside the computing nodes, but tightly coupled with them. These agents monitor the power consumption of each computing node at the plug as well as performance and utilization metrics. The monitored values are sent to a data management backbone, through a communication layer based on the open-source MQTT protocol, a TCP/IP protocol designed for low bandwidth and high latency networks, with minimal resource demands.

Since there are limitations on the storage space available for the monitoring infrastructure, it is impossible to store the raw data. The solution adopted in \davide~was to discard the fine-grained data older than a week and to preserve indefinitely job information and coarse-grained data (long term storage, around 6GB after 7 months of activity). For this paper, we work with the coarse-grained data aggregated in 5-minutes long intervals. Furthermore, we focused on a subset of the data collected by Examon; for each node we have 166 metrics (our \emph{features}), i.e. core loads, temperatures, fan speed, power consumptions, etc.

\section{The Autoencoder-based Approach}
\label{sec:ae_model}

We aim at detecting anomalies that happen at the node-level. Currently, we focus on single nodes. We create a set of separate autoencoder models, one for each node in the system. Each model is trained to learn the normal behaviour of the corresponding node and to be activated if anomalous conditions are measured.
If an autoencoder can learn the correlations between the set of measurements (features) that describe the state of a supercomputer, then it can consequently notice changes in these correlations that indicate an abnormal state. Under normal operating conditions these features are linked by specific relations (i.e. the power consumption of a core is directly related to the workload and temperature to the power and frequency). We hypothesize that these correlations will be perturbed if the system enters in an anomalous state. 

The \emph{reconstruction error} is the element we use to detect anomalies. An autoencoder can be trained to minimize this error. In doing so, it learns the relationships among the features of the input set. If we feed a trained autoencoder with data not seen during the training phase, it should reproduce the new input with good fidelity, at least if the new data resemble the data used for the training. If this is not the case, the autoencoder cannot correctly reconstruct the input and the error will be greater. We propose to detect anomalies by observing the magnitude of the reconstruction error.

All autoencoders have the same structure. We opted for a fairly simple structure composed by three layers: I) an input layer with as many neurons as the number of features (166), II) a densely connected intermediate sparse layer \cite{boureau2008sparse} with 1660 neurons (ten times the number of features) with Rectified Linear Units (\emph{ReLu}) as activation functions and a L1 norm regularizer \cite{goodfellow2016deep}, III) a final dense output layer with 166 neurons with linear activations. This network was obtained after an empirical evaluation, after having experimented with different topologies and parameter configurations. 
To summarize, our methodology has the following steps: 1) create an autoencoder for each computing node in the supercomputer; 2) train the autoencoders using data collected during normal operating conditions; 3) identify anomalies in new data using the reconstruction error obtained by the autoencoders.

\section{Experimental Evaluation}
\label{sec:exps}

In every HPC system there are multiple possible sources of anomalies and fault conditions, ranging from hardware faults to software errors. In this paper we verify the proposed approach on a type of anomaly that easily arises in real systems and happens at the level of single nodes, namely \emph{misconfiguration}. More precisely, we consider the misconfiguration of the frequency governor of a computing node. Modern Linux systems allow to specify different policies regulating the clock speed of the CPUs, thanks to kernel-level drivers referred as frequency governors \cite{brodowski2013cpu}. Different policies have different impacts on the clock speed, frequency and power consumption of the CPUs.

We considered three different policies. The first one, \emph{conservative}, is the default policy on \davide~(the normal behaviour); it sets the CPU clock depending on the current CPU load. Two other types of policies have been used to generate anomalies, i) the \emph{powersave} policy and ii) the \emph{performance} policy. These frequency governors statically set the CPU to the, respectively, lowest and highest frequency in the allowed range.

\subsection{Results}
\label{sec:results}
In this work we used an off-line approach. We gathered the measurements collected during months of real usage of \davide~and we created a data set; the data is normalized to have values in the range $[0,1]$. The data set is split in 3 components: 1) the training set $D_{Train}$ (containing data points within periods of normal behaviour), 2) the test set without anomalies $D_{Test}^N$ (again, only periods of normal behaviour) and 3) the test set with anomalies $D_{Test}^A$ (the periods when we injected anomalies on some nodes).

For these experiments we selected a subset of the data collected by Examon during \davide~lifetime. The period we considered is 83 days long, from March 2018 to May 2018. During this period \davide~was in the normal state for most of the time -- 66 days, ~80\% of the time -- while we forced anomalous states for smaller sub-periods of a few days, 13 days in total. Since we know when the anomalies were injected identifying $D_{Test}^A$ is trivial. $D_{Train}$ and $D_{Test}^N$ were created by randomly splitting the data points belonging to the 66 days of normal state, 80\% of the data points going to $D_{Train}$ and 20\% to $D_{Test}^N$. 

Each autoencoder is trained with \emph{Adam} \cite{kingma2014adam} optimizer with standard parameters, minimizing the mean absolute error; the number of epochs used in the training phase is 100 and the batch size has a fixed value (32). These values were chosen after a preliminary exploration because they guarantee very good results with very low computational costs. The time required to train the network is around 5 minutes on a quad-core processor (Intel i7-5500U CPU 2.40GHz) with 16GB of RAM (without using GPUs).

\subsubsection{Reconstruction Error-Based Detection}
\label{sec:results_error}

As explained previously, our anomaly detection method relies on the hypothesis that an autoencoder can be taught to learn the correlations among the features in a data set representing the healthy state of a supercomputer node. In this case the autoencoder would be capable to reconstruct an input data set never seen before, if this new input resembles the healthy one used during the training phase -- if in the unseen data set the features correlations are preserved. Conversely, an autoencoder would struggle to reconstruct data sets where the learned correlations do not hold. To demonstrate our hypothesis, we expect to observe higher reconstruction errors for the anomalous periods with respect to the error obtained in normal periods. We are not strictly interested in the absolute value of the reconstruction error but rather on the relative difference between normal and anomalous periods.

\begin{figure}[hbt]
	\centering
	\includegraphics[width=0.45\textwidth,height=\figH]{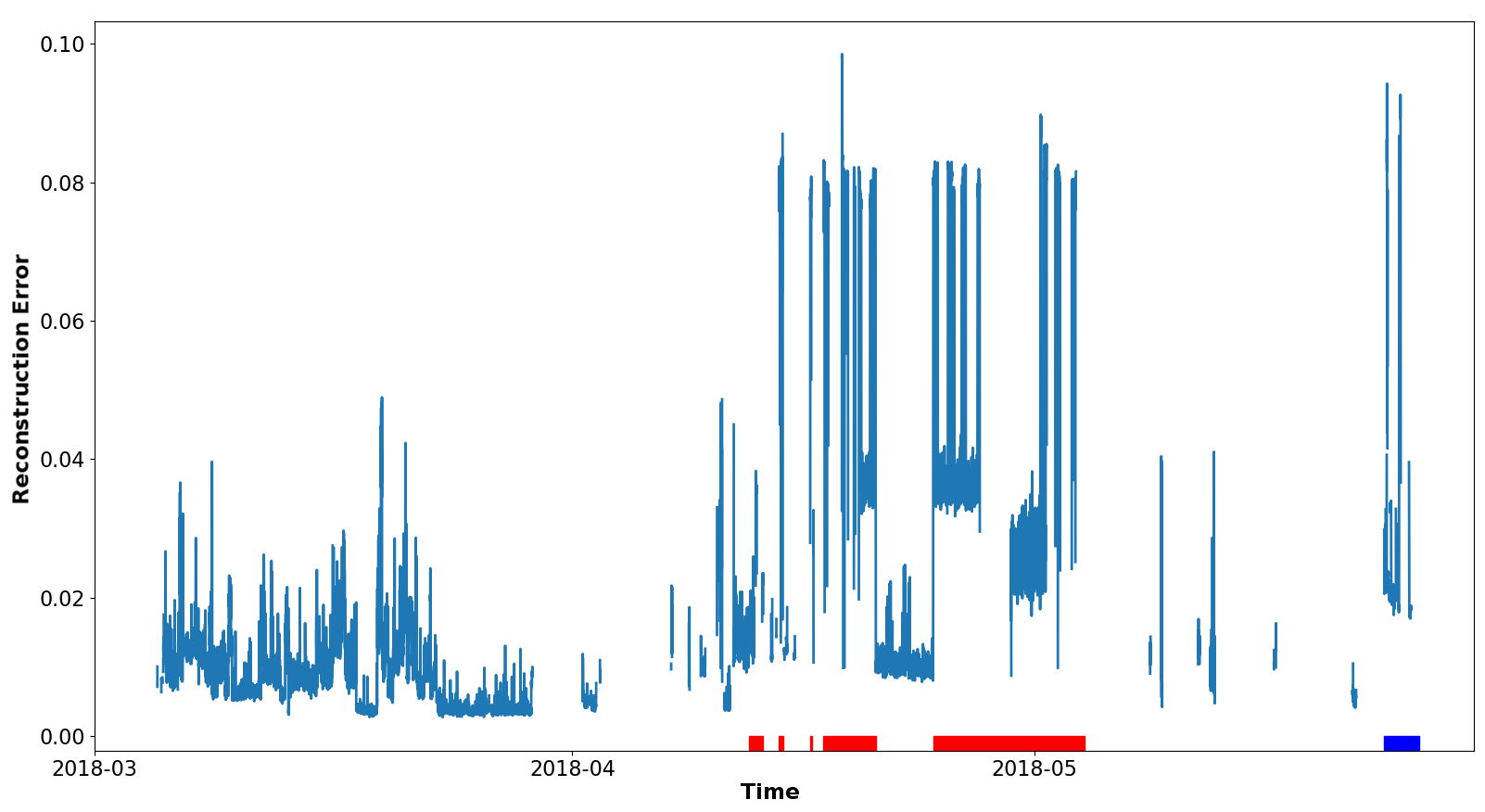}
	\caption{Reconstruction error for node \emph{davide45}}
	\label{fig:recon_error_nodes_45}
\end{figure}

This reconstruction error is plotted in Figure~\ref{fig:recon_error_nodes_45}; it displays the results computed for node \emph{davide45} (other nodes were omitted for space reason but their behaviour is very similar). The $x$-axis and $y$-axis show, respectively, the time and the normalized reconstruction error (we sum the error for each feature and divide by the number of features $NF$). The  reconstruction error trend is plotted with a light blue line; the gaps in the line represent periods when the node was idle and that have been removed from the data set. 

We observe 6 anomalous periods (highlighted by colored lines along the $x$-axis): during the first 5 (red lines) the frequency governor was set to powersave while during the last one (blue) the governor was set to performance. The reconstruction error is never exactly zero, but this is not our concern: our analysis does not rely on the absolute value of the error, but rather on the relative magnitude of the errors computed for different data sets.
The reconstruction error is indeed greater when the nodes are in an anomalous state, as underlined by the higher values in the $y$-axis in the periods corresponding to anomalies. Hence, the autoencoder struggles to recreate the ``faulty'' input data set. 

Although the plot shown is promising, it does not actually show that the reconstruction error for unseen healthy input is actually lower than the error committed with anomalous periods. This happens because the normal behaviour data set was randomly split in the subset $D_{Train}$ and $D_{Test}^N$ and it is impossible to distinguish between them by simply looking at the plot. However, our insight is backed by the quantitative analysis, as summarized in Table~\ref{tab:quantitative_analysis}. To measure the quality of the anomaly detection we rely on the Mean Absolute Error (MAE) and on the Root Mean Squared Error (RSME). For each autoencoder we computed MAE and RSME for every set $D_{Train}$, $D_{Test}^N$ and $D_{Test}^A$.

The results obtained for all autoencoders are very similar but in order to make a fair comparison between different nodes we do not use the absolute values of MAE and RSME but we rather employ a normalized version: the normalized MAE (RSME) is obtained by dividing the actual MAE (RSME) by the MAE (RSME) computed for $D_{Train}$. In this way we force the normalized error for the training set to be equal to 1 (since we are not strictly interested in its absolute value) and we highlight the relative difference of error between sets. If the normalized error for a test set is close to one this means that the autoencoder was able to reconstruct the input quite well; larger errors imply that the autoencoder was not capable to reproduce the input -- these situations are those that we claim to be anomalies.

In Table~\ref{tab:quantitative_analysis} we can see the results for the test sets. The error for the $D_{Test}^A$ is the average value obtained considering both anomaly types. The normalized error for $D_{Train}$ has not been reported since it is always equal to 1. The values reported in the table are the average computed over all the autoencoders (as many as in the chosen subset of nodes of \davide).
\begin{table}
\small\sf\centering
\begin{tabular}{lrr}
 \toprule
	 \qquad& $D_{Test}^N$ \qquad& $D_{Test}^A$ \\ 
  \midrule
Normalized MAE \qquad& 1.08 \qquad& 14.54 \\
Normalized RMSE \qquad& 1.17 \qquad& 11.18 \\
  \bottomrule
\end{tabular}
\caption{Quantitative Analysis: average over all nodes}
\label{tab:quantitative_analysis}	
\end{table}
The results clearly indicate that our hypothesis holds true (as hinted also by the previous plot with the reconstruction error). Both the average normalized MAE and RSME for the test set with no anomalies $D_{Test}^N$ are very close to 1, suggesting that the autoencoders have correctly learned the correlations between the measured features of a healthy system. Therefore, when the autoencoders are fed with unseen input that preserve these correlations they can reconstruct it with good precision. On the contrary, the autoencoders cannot correctly reproduce new input that does not resemble a healthy system, that is a system in an anomalous state. This is shown by the markedly higher normalized MAE and RSME obtained for $D_{Test}^A$. 

We can also make an additional observation: there is clearly a difference between the reconstruction errors of the two anomalies (powersave and performance), as highlighted by the notably different values of MAE and RSME for $D_{Test}^{A1}$ and $D_{Test}^{A2}$. In particular, larger normalized MAE and RMSE suggest that the autoencoders are less capable to reconstruct the input composed by data points with powersave frequency governor. In a certain sense, this means that this frequency governor policy is a more anomalous condition than the performance case, i.e. it disrupts the features correlations in a more marked way. Why does this happens? A possible explanation is that the performance frequency governor has a behaviour closer to the default conservative one: when a computing node in a supercomputer is active, its frequency tends to reach the maximum allowed value quite rapidly because HPC applications are optimized to optimally use the computing resources at their disposal. Thus, the performance mode has a more similar footprint to the conservative one compared to the powersave case, where instead the frequency is fixed to the lowest possible value, not a common situation for an active HPC node.

\subsubsection{Detection Accuracy}
\label{sec:results_classify}

So far we have observed the reconstruction error trends obtained by our approach based on autoencoders, but we still have to discuss how the reconstruction error can be used to actually detect an anomaly. Our goal is to identify an error threshold $\theta$ to discriminate between normal and anomalous behaviour. In order to do so we shall start by looking at the distributions of the reconstruction errors. Again, we are considering each autoencoder (and thus corresponding node) separately. We distinguish the errors distribution for healthy data sets ($D_{Train} \cup D_{Test}^N$) and for the unhealthy data set ($D_{Test}^A$). 

\begin{figure*}[!t]
        \centering
	\subfloat[Normal data set]{
		\includegraphics[width=0.47\textwidth,height=\figHsmall]{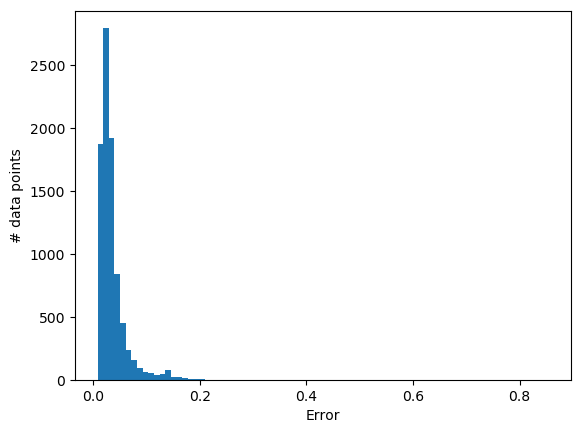}
		\label{fig:noAnomalies_maxError_distribution_node45_20180303_20180526}
	}%
	\hfill
	\subfloat[Anomaly data set]{
		\includegraphics[width=0.47\textwidth,height=\figHsmall]{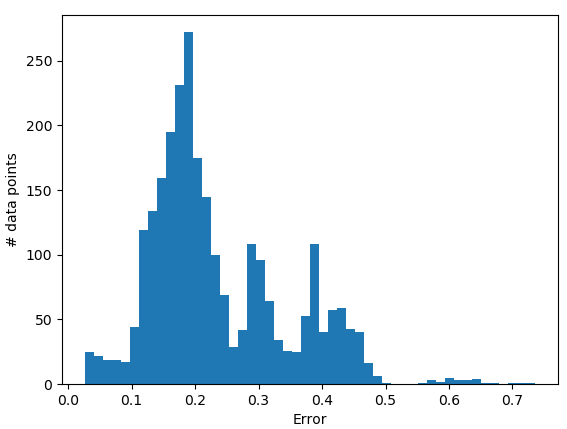}
		\label{fig:anomalies_maxError_distribution_node45_20180303_20180526}
	}%
\caption{Error distribution for node \emph{davide45}}
\label{fig:error_distrib_45}
\end{figure*}

Figure~\ref{fig:error_distrib_45} shows the error distributions for the autoencoder corresponding to node \emph{davide45} -- again other nodes have the same behaviour. The graph contains the histograms of the error distributions; in the $x$-axis we have the reconstruction error and in the $y$-axis there is the number of data points with the corresponding error. The left-most sub-figure (Fig.~\ref{fig:noAnomalies_maxError_distribution_node45_20180303_20180526}) shows the error distribution for the normal data set ($D_{Train} \cup D_{Test}^N$) and the other one (Fig.~\ref{fig:anomalies_maxError_distribution_node45_20180303_20180526}) shows the distribution for the anomalous data set. It is quite easy to see that the errors distribution of the normal data set is extremely different from the anomalous one. 

Since we can clearly distinguish the error distributions we opted for a simple method to classify each data point: if the reconstruction error $E_i$ for data point $i$ is greater than a threshold $\theta$, then the point is ``abnormal''; otherwise the data point is considered normal. The next step is to identify the threshold used to classify each data point. We choose as a threshold the $n$-th percentile of the errors distribution of the normal data set, where $n$ is a value that depends on the specific autoencoder/node. For example in the case of \emph{davide45} (Fig.~\ref{fig:error_distrib_45}), if $n=95$ the threshold is equal to 0.082; this means that 95\% of the errors in the normal data set (Fig.~\ref{fig:noAnomalies_maxError_distribution_node45_20180303_20180526}) are smaller than this value. Hence, if a data point fed to our trained autoencoder generates an error greater than 0.082 we classify it as anomalous. In order to find the best $n$ value for each autoencoder we employed a simple generate-and-test search strategy, that is we performed experiments with a finite number of values and then chose those guaranteeing the best results in term of classification accuracy. 

Broadly speaking, the best results are obtained with higher thresholds, i.e. $n \geq 93$. To asses the accuracy of the classification (thus the goodness of the threshold) we compute the \emph{F-score} \cite{van1979information} for each class; in our case we have two classes, \emph{normal} (N) and \emph{anomaly} (A). The F-score is a widespread metric to measure the accuracy of a classification test and can assume values in the $[0,1]$ range, with values closer to 1 indicating a higher accuracy. In Table~\ref{tab:classification_accuracy} we see some results. In the first column from the left there is the node whose autoencoder F-score values are reported (we report the values for only a subgroup of nodes). The remaining columns report the F-score values for 3 different $n$-th percentiles (and therefore different thresholds); there are two F-score values for each $n$-th percentile, one computed for the normal class (\emph{N}) and one for the anomaly class (\emph{A}). 

\begin{table}
\small\sf\centering
\begin{tabular}{ccccccc}
 \toprule
 \multirow{2}{*}{\emph{Node}} & \multicolumn{2}{c}{95-th perc.} & \multicolumn{2}{c}{97-th perc.} & \multicolumn{2}{c}{99-th perc.} \\
  & N & A & N & A & N & A \\ 
  \midrule
\emph{davide17} & 0.97 & 0.89 & 0.98 & 0.93 & 0.99 & 0.97 \\  
\emph{davide19} & 0.97 & 0.90 & 0.98 & 0.94 & 0.99 & 0.97 \\  
\emph{davide45} & 0.97 & 0.92 & 0.98 & 0.95 & 0.99 & 0.98 \\ 
  \midrule
\emph{davide27} & 0.95 & 0.90 & 0.91 & 0.77 & 0.86 & 0.52 \\  
\emph{davide28} & 0.94 & 0.88 & 0.96 & 0.89 & 0.90 & 0.69 \\   
\emph{davide29} & 0.97 & 0.75 & 0.98 & 0.82 & 0.99 & 0.85 \\  
  \midrule
  \midrule
\emph{Average} & 0.96 & 0.87 & 0.96 & 0.88 & 0.95 & 0.82 \\
  \bottomrule
\end{tabular}
\caption{Classification Results}
\label{tab:classification_accuracy}	
\end{table}

The table can be divided in three subparts (separated by horizontal lines): 1) the first one contains nodes similar to \emph{davide45}, i.e. nodes where most of the anomalies were of type powersave; 2) the second group is comprised of nodes where most of the anomalies had the frequency governor set to performance; 3) the last group (the last row) is the average of the other nodes. In general we can see that the F-score values are very good, highlighting the high accuracy of our approach. A notable difference can be observed between the two sub-groups of nodes. In nodes with a prevalence of powersave anomalies higher thresholds (higher $n$-th values) guarantee better results: this happens because, as seen for instance in Figure~\ref{fig:error_distrib_45}, the error distributions are more separable. In the case of nodes characterized by more anomalies of performance type, increasing the threshold does not necessarily improve the accuracy -- although this can still occur for some nodes. In these nodes it is harder to distinguish normal data points from anomalies of type performance (since they behave similarly). Hence, simply increasing the threshold is not beneficial, for example for both \emph{davide27} and \emph{davide28} the best $n$-th value found through our empirical exploration is $n = 94$. The underlying reason is that increasing the threshold (for these nodes) leads to a marked increase in the number of false negatives.   

\section{Conclusion}
\label{Conclusion}

In this paper we proposed an approach to detect anomalies in a HPC system that relies on large data sets collected via a lightweight and scalable monitoring framework and employs autoencoders to distinguish between normal and anomalous system states.

In the future we plan to further validate our method by testing it on a broader set of anomalies. Our goal is to expand the anomaly detection technique in order to be able to also classify different types of anomalies; in addition to recognize that the system is in an anomalous state, the autoencoder (possibly a refined and more complex version) will be also able to distinguish among different anomaly classes and sources.
We also plan to implement our approach in a on-line prototype to perform real-time anomalous detection on a supercomputer, again using \davide~as a test bed.

\subsection*{\textbf{Acknowledgements}}
\noindent
This work was partially supported by the FP7 ERC Advanced project MULTITHERMAN (g.a. 291125).
We also want to thank CINECA and E4 for granting us the access to their systems.

\bibliographystyle{alpha}
\bibliography{bib}

\end{document}